\documentclass[letterpaper]{article}
\PassOptionsToPackage{table}{xcolor}

\usepackage[preprint]{kosmo}
\usepackage{ragged2e}
\usepackage{graphicx}
\usepackage{amsmath}
\usepackage{amssymb}
\usepackage{booktabs}
\usepackage{multirow}
\usepackage{xcolor}
\usepackage{caption}

\title{GeoWeaver: Accurate Long-Sequence 3D Reconstruction via Hierarchical Geometric Assembly}

\author{
  Tinghao Jiang\textsuperscript{1,2},
  Sheng Tang\textsuperscript{1},
  Shengzhe Wei\textsuperscript{1},
  Juntong Fang\textsuperscript{3},\\
  Weiqi Zhang\textsuperscript{3},
  Junsheng Zhou\textsuperscript{3},
  Zesong Li\textsuperscript{1}
}

\affiliations{
  \textsuperscript{1}Kosmo Research\quad
  \textsuperscript{2}Shanghai Jiao Tong University\quad
  \textsuperscript{3}Tsinghua University
}

\correspondingauthor{}

\projectpage{\url{http://kosmoresearch.github.io/GeoWeaver/}}

\kosmoabstract{
\justifying
Long-sequence 3D reconstruction from RGB videos requires both accurate local
geometry and globally consistent camera motion. Feed-forward models provide
strong depth and pose predictions, but their memory cost prevents joint
inference over long sequences. Chunk-wise processing improves scalability,
yet independently predicted chunks often exhibit scale drift, pose errors,
and point-cloud misalignment.
We present \textbf{GeoWeaver}, a unified framework comprising a
\textit{Geometric Prior Model} (GPM) and \textit{Test-Time Adaptation} (TTA).
The GPM predicts chunk-wise depth, confidence, and camera parameters as
adjustable geometric priors. TTA then performs sequential initialization,
global chunk-level $\mathrm{Sim}(3)$ alignment, and coarse-to-fine refinement
of camera poses, affine depth corrections, and intrinsics. Dense
correspondences provide adjacent, cross-chunk, and long-range constraints,
while a robust CDF-style objective jointly optimizes weighted 2D reprojection
and 3D consistency residuals. This design preserves local geometric accuracy
while correcting accumulated pose, scale, depth, and calibration errors.
Experiments across diverse long-sequence benchmarks demonstrate improved
camera accuracy, global consistency, and point-cloud quality. Ablations verify
the contribution of each adaptation stage, and applying the same TTA procedure
to different geometric prior models consistently improves their trajectory
estimates, demonstrating that GeoWeaver is not tied to a specific GPM.
}

\providecommand{\catopt}{Opt.}
\providecommand{\catff}{FF}
\providecommand{\catchunk}{Chunk-FF}
\providecommand{\catstream}{Stream}
\providecommand{\cathybrid}{Hybrid}
\providecommand{\tablefontsize}{\footnotesize}
\providecommand{\widetablefontsize}{\scriptsize}
\providecommand{\densetablefontsize}{\fontsize{6pt}{6.9pt}\selectfont}
\providecommand{\widetablerowstretch}{1.06}
\providecommand{\tabcite}[1]{[\citenum{#1}]}

\providecommand{\rankone}[1]{\textbf{#1}}
\providecommand{\ranktwo}[1]{\underline{#1}}
\providecommand{\rankthree}[1]{#1}
\providecommand{\ourscell}[1]{\textbf{#1}}
\providecommand{\oursval}[1]{#1}
\providecommand{\oursrankone}[1]{\textbf{#1}}
\providecommand{\oursranktwo}[1]{\underline{#1}}
\providecommand{\oursrankthree}[1]{#1}

\newcommand{\suppsingletablestyle}{%
  \footnotesize
  \setlength{\tabcolsep}{3.0pt}%
  \renewcommand{\arraystretch}{1.08}%
}

\begin{document}

\maketitle

\section{Introduction}
\label{sec:introduction}

Long-sequence 3D reconstruction recovers camera motion and scene geometry from
temporally ordered RGB images, supporting autonomous driving, robotics,
augmented reality, and large-scale scene understanding. Compared with short
sequences, long videos involve larger motion, revisited regions, illumination
changes, weak overlap, and accumulated drift.

Classical SfM and SLAM enforce global consistency through matching,
registration, triangulation, pose estimation, and bundle
adjustment~\cite{colmap,glomap,orbslam3,dso}. They perform well with reliable
correspondences and well-conditioned view graphs, but often degrade under weak
texture, repetitive structures, limited overlap, or long trajectories.

Feed-forward geometry models provide a complementary solution.
DUSt3R~\cite{dust3r}, MASt3R~\cite{mast3r}, VGGT~\cite{vggt}, and
Depth Anything~3~\cite{da3} directly predict depth, cameras, point maps,
tracks, or multi-view geometry. Although they provide strong learned priors,
their memory and attention costs make joint inference over hundreds or
thousands of frames impractical.

\begin{figure}[!t]
\centering
\includegraphics[
    width=\columnwidth,
    keepaspectratio
]{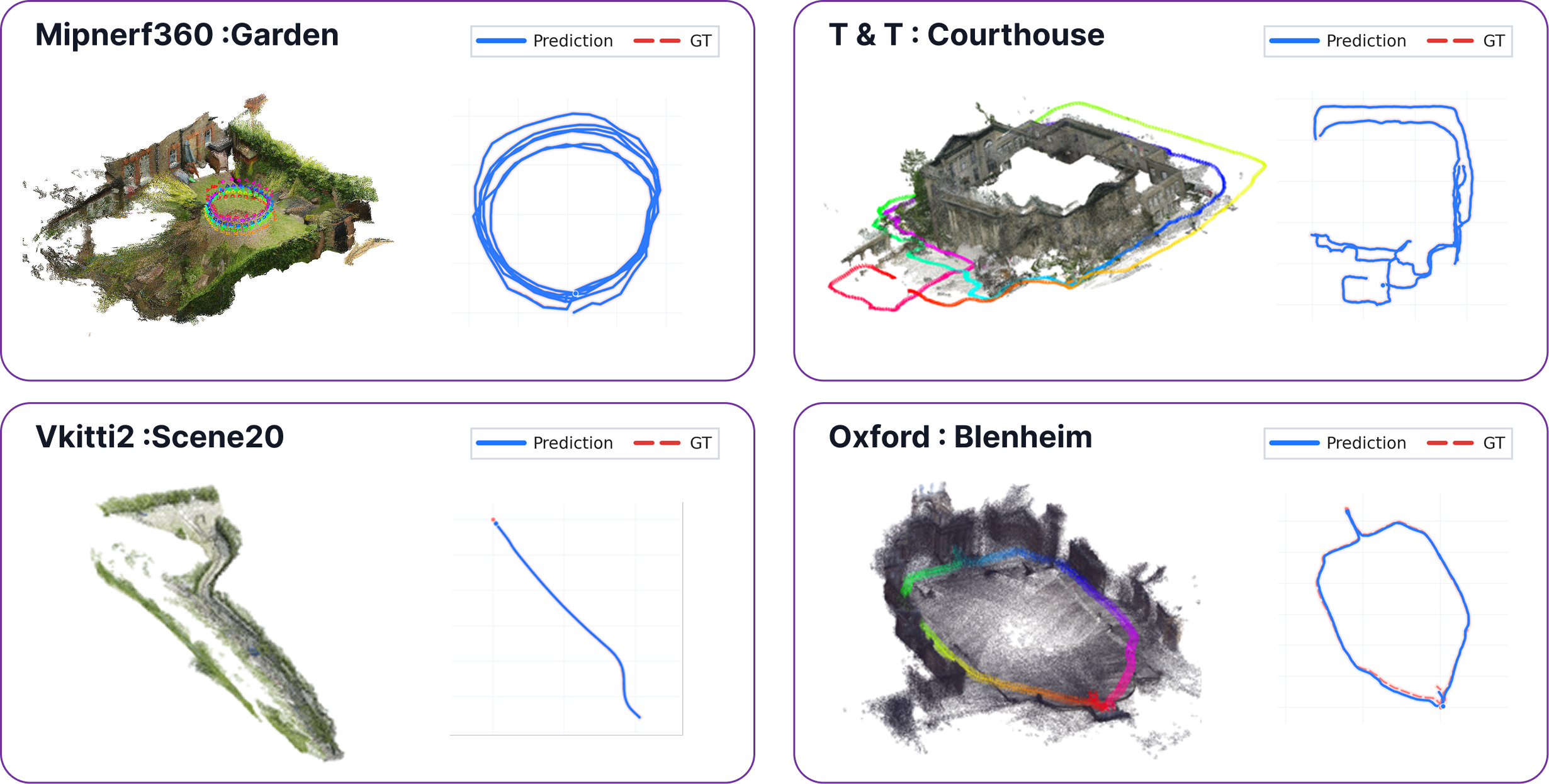}
\caption{
\textbf{Qualitative results across four benchmarks.}
Reconstructed point clouds and camera trajectories are shown for each scene.
}
\label{fig:first_page_qualitative}
\end{figure}

A common solution is chunk-wise inference, which reconstructs short windows
independently and then assembles them. While bounding inference cost, this
introduces inconsistent coordinate frames, scales, and calibration across
chunks. Simple overlap alignment leaves local errors largely unchanged,
whereas directly optimizing all frame-level variables before establishing a
stable global layout is prone to poor local minima. The key challenge is thus
to preserve reliable local priors while correcting sequence-level drift.

We address this challenge with \textbf{GeoWeaver}, which combines a Geometric
Prior Model (GPM) with sequence-specific Test-Time Adaptation (TTA). The GPM
predicts chunk-wise depth, confidence, and camera parameters. TTA sequentially
initializes the chunk layout, corrects global scale and pose inconsistencies
through chunk-level $\mathrm{Sim}(3)$ optimization, and refines frame poses,
affine depth, and camera-group focal corrections. Confidence-weighted 2D and
3D constraints couple all stages under a robust CDF-style objective.
Figure~\ref{fig:first_page_qualitative} shows the resulting coherent geometry
and trajectories.

Our main contributions are:

\begin{itemize}

\item We introduce a feed-forward Geometric Prior Model that predicts depth,
confidence, and camera parameters from variable-length multi-view inputs.

\item We propose a minimal-overlap TTA strategy that converts independent
chunk priors into a stable global layout.

\item We develop a hierarchical TTA procedure combining chunk-level
$\mathrm{Sim}(3)$ alignment with joint refinement of frame poses, affine
depth, and camera-group focal corrections under robust 2D and 3D constraints.

\item We demonstrate consistent gains across diverse long-sequence benchmarks
and show that the same TTA procedure improves priors from multiple
feed-forward models without source-specific tuning.

\end{itemize}

\section{Related Work}
\label{sec:related_work}

\begin{figure*}[t]
\centering
\includegraphics[width=\textwidth]{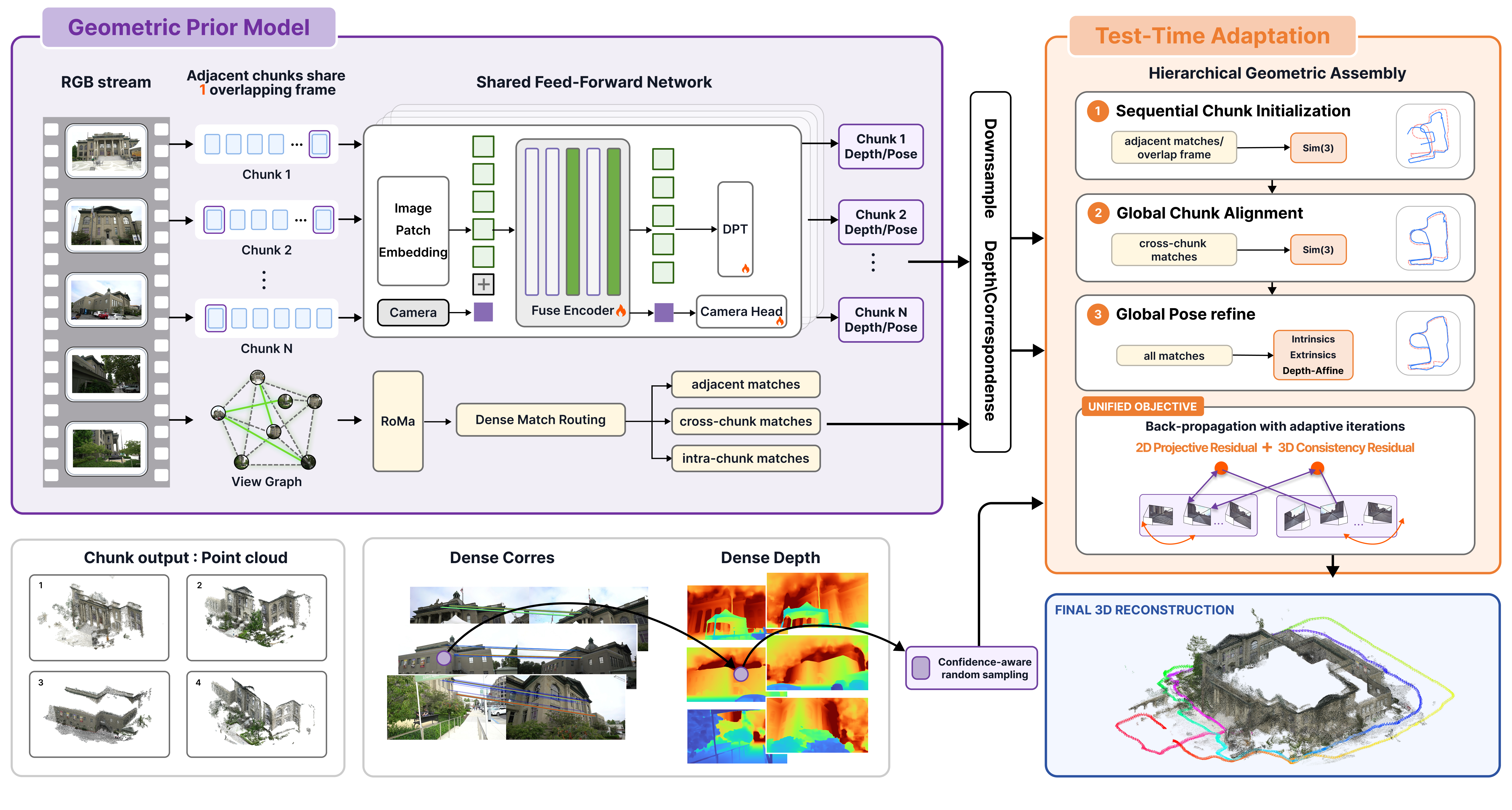}
\caption{
Overview of GeoWeaver. The Geometric Prior Model predicts chunk-wise depth,
confidence, and cameras, while Test-Time Adaptation performs sequential
initialization, chunk alignment, and frame-level refinement using local and
long-range correspondences.
}
\label{fig:overview}
\end{figure*}

\paragraph{Optimization-based reconstruction.}
Classical SfM and SLAM recover cameras and geometry through matching,
registration, triangulation, and bundle adjustment. COLMAP~\cite{colmap}
uses incremental SfM, GLOMAP~\cite{glomap} estimates globally consistent
cameras, and ORB-SLAM3~\cite{orbslam3}, DSO~\cite{dso}, and
DROID-SLAM~\cite{droid} exploit temporal continuity for tracking and mapping.
Detector-Free SfM~\cite{dfsfm} further combines dense matching with iterative
track and geometry refinement. Despite strong global accuracy, these methods
remain sensitive to weak texture, repeated structure, motion blur, limited
overlap, and incorrect associations, and do not directly adapt dense learned
geometric priors.

\paragraph{Feed-forward geometry models.}
Feed-forward methods infer multi-view geometry directly from RGB images.
DUSt3R~\cite{dust3r} predicts dense point maps, while
MASt3R~\cite{mast3r} augments them with dense matching. VGGT~\cite{vggt},
Depth Anything~3~\cite{da3}, $\pi^3$~\cite{pi3}, Pow3R~\cite{pow3r}, and
MapAnything~\cite{mapanything} predict depth, cameras, point maps, or metric
geometry from uncalibrated views. Fast3R~\cite{fast3r} and
Speed3R~\cite{speed3r} improve scalability, while
Light3R-SfM~\cite{light3rsfm} and SAIL-Recon~\cite{sailrecon} extend
feed-forward SfM through global alignment or localization. These models provide
strong local priors, but joint inference over very long sequences remains
memory-intensive.

\paragraph{Long-sequence and chunk-wise reconstruction.}
Long-sequence methods scale through persistent memory, causal state
propagation, incremental registration, or chunk-wise inference
~\cite{spann3r,cut3r,long3r,stream3r,streamvggt,lingbotmap}.
Spann3R~\cite{spann3r} and CUT3R~\cite{cut3r} maintain persistent scene
representations; LONG3R~\cite{long3r}, STream3R~\cite{stream3r},
StreamVGGT~\cite{streamvggt}, and WinT3R~\cite{wint3r} process image streams
causally or within windows. SLAM3R~\cite{slam3r} and
LingBot-Map~\cite{lingbotmap} incrementally register or memorize geometric
context, while VGGT-Long~\cite{vggtlong}, Scal3R~\cite{scal3r},
ZipMap~\cite{zipmap}, Online3R~\cite{online3r}, and LoGeR~\cite{loger}
improve long-range consistency through overlap, alignment, adaptation, or
memory. However, local depth, pose, scale, and calibration errors often remain
fixed or only weakly adjustable.

\paragraph{Hybrid learning and geometric optimization.}
Hybrid methods combine learned geometry with explicit optimization.
BA-Net~\cite{banet} introduces differentiable feature-metric bundle
adjustment, DROID-SLAM~\cite{droid} jointly updates poses and depth, and
FlowMap~\cite{flowmap} optimizes depth, intrinsics, and cameras from
correspondences. VGGSfM~\cite{vggsfm}, MASt3R-SfM~\cite{mast3rsfm},
MASt3R-SLAM~\cite{mast3rslam}, VGGT-SLAM~\cite{vggtslam}, and
VGGT-SLAM~2.0~\cite{vggtslam2} integrate learned tracking, matching, or local
reconstructions into global optimization. MP-SfM~\cite{mpsfm},
Marginalized Bundle Adjustment~\cite{mba}, and AMB3R~\cite{amb3r} further
combine learned geometric priors with explicit reconstruction.

GeoWeaver follows this hybrid direction but targets minimal-overlap
long-sequence reconstruction. It treats chunk-wise depth, confidence, and
camera estimates as adjustable priors and adapts them to each sequence through
hierarchical geometric optimization.

\begin{figure*}[t]
\centering
\includegraphics[width=\textwidth]{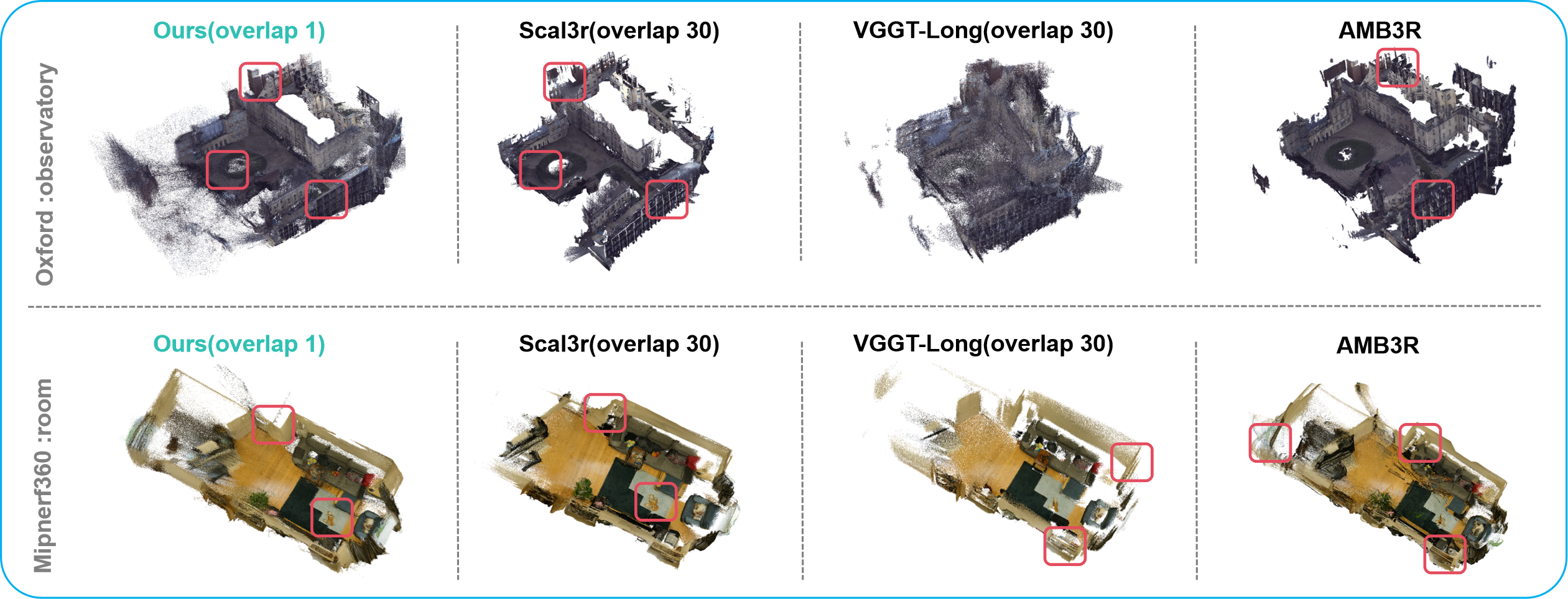}
\caption{
Qualitative comparison on representative indoor and outdoor scenes.
``Overlap'' denotes the number of shared frames between adjacent 60-frame
chunks. Compared with baselines using 30-frame overlap, GeoWeaver uses only
one shared frame while producing more coherent geometry, better aligned
trajectories, and fewer inter-chunk discontinuities.
}
\label{fig:qualitative}
\end{figure*}

\section{Method}
\label{sec:method}

As shown in Fig.~\ref{fig:overview}, GeoWeaver comprises a Geometric Prior
Model (GPM) and Test-Time Adaptation (TTA). The GPM independently predicts
depth, confidence, and camera parameters for short overlapping chunks. With
the GPM fixed, TTA first initializes the chunk layout from adjacent
correspondences, then optimizes one $\mathrm{Sim}(3)$ transformation per chunk
using cross-chunk and long-range constraints, and finally refines frame poses,
affine depth, and camera-group focal corrections. This coarse-to-fine process
corrects global scale and coordinate inconsistencies before frame-level
refinement.

\subsection{Geometric Prior Model}
\label{sec:chunk_priors}

We split a long RGB sequence into contiguous chunks with one shared frame
between adjacent chunks, providing a geometric anchor with minimal redundant
inference. A shared physical frame yields two chunk-specific observations; in
Stages~1--2, they are treated independently and $c(i)$ denotes the chunk of
observation $i$. After chunk alignment, Stage~3 retains the observation with
larger mean valid confidence as the canonical prediction; the other is used
only to constrain chunk alignment.

For $\mathcal{C}_k=\{I_i\}_{i=1}^{N_c}$, the Geometric Prior Model (GPM)
predicts
\begin{equation}
\begin{aligned}
\mathcal{G}_i
&=
\{
\hat{D}_i,
\hat{C}_i,
\hat{\boldsymbol{\theta}}_i
\},\\
\hat{\boldsymbol{\theta}}_i
&=
(
\hat{\mathbf{t}}_i,
\hat{\mathbf{q}}_i,
\widehat{\mathrm{FoV}}_i
).
\end{aligned}
\label{eq:local_priors}
\end{equation}
Here, $\hat{D}_i$, $\hat{C}_i$, and
$\hat{\boldsymbol{\theta}}_i$ denote depth, confidence, and camera parameters,
from which poses and intrinsics are recovered.

Built on Depth Anything~3~\cite{da3}, the GPM uses a ViT encoder, a
DPT-style depth-confidence head, and a transformer camera head. It is trained
with joint geometric and camera supervision, progressing from fixed four-view
clips to variable-length clips of 2--16 views.

The geometric model loss combines local geometry, normalized chunk-level geometry,
and relative camera supervision:
\begin{equation}
\begin{aligned}
\mathcal{L}_{\mathrm{front}}
=
&
\lambda_{\mathrm{local}}
\mathcal{L}_{\mathrm{local}}
+
\lambda_{\mathrm{global}}
\mathcal{L}_{\mathrm{global}}
+
\lambda_{\mathrm{cam}}
\mathcal{L}_{\mathrm{cam}} .
\end{aligned}
\label{eq:front_loss}
\end{equation}
The local term operates in each camera coordinate system, whereas the global
term transforms predictions with the estimated cameras and enforces consistency
in a normalized chunk coordinate system. Both use
\begin{equation}
\begin{aligned}
\mathcal{L}_{\mathrm{geo}}
=
\frac{1}{|\Omega|}
\sum_{\mathbf{u}\in\Omega}
M(\mathbf{u})
\Big[
&
\hat{C}(\mathbf{u})
\|
\hat{\mathbf{X}}(\mathbf{u})
-
\mathbf{X}(\mathbf{u})
\|_1
\\
&
-\alpha\log \hat{C}(\mathbf{u})
\Big],
\end{aligned}
\label{eq:geo_loss}
\end{equation}
where $\Omega$ is the image domain, $M$ is a valid-pixel mask, and
$\hat{\mathbf{X}}$ and $\mathbf{X}$ are predicted and ground-truth 3D points.
The strictly positive confidence $\hat{C}$ weights geometric error, while its
logarithmic term prevents trivial confidence suppression.

Camera supervision uses relative transformations:
\begin{equation}
\begin{aligned}
\hat{\mathbf{T}}_{ij}
&=
\hat{\mathbf{T}}_j
\hat{\mathbf{T}}_i^{-1},
\\
\mathbf{T}^{*}_{ij}
&=
\mathbf{T}^{*}_j
(\mathbf{T}^{*}_i)^{-1}.
\end{aligned}
\label{eq:relative_pose}
\end{equation}
A shared chunk-level scale $s^*$, estimated from depth-induced geometry,
resolves monocular translation ambiguity:
\begin{equation}
\begin{aligned}
\mathcal{L}_{\mathrm{cam}}
=
\frac{1}{N_c(N_c-1)}
\sum_{i\neq j}
\Big[
&
\ell_{\mathrm{rot}}
(
\hat{\mathbf{R}}_{ij},
\mathbf{R}^{*}_{ij}
)
\\
&
+
\lambda_{\mathrm{trans}}
\ell_{\mathrm{trans}}
(
s^*\hat{\mathbf{t}}_{ij},
\mathbf{t}^{*}_{ij}
)
\Big].
\end{aligned}
\label{eq:camera_loss}
\end{equation}
We use geodesic rotation error and a Huber translation penalty:
\begin{equation}
\ell_{\mathrm{rot}}
=
\arccos
\left[
\operatorname{clip}
\left(
\frac{
\operatorname{tr}
(
\hat{\mathbf{R}}_{ij}^{\top}
\mathbf{R}^{*}_{ij}
)-1
}{2},
-1,1
\right)
\right],
\label{eq:rotation_loss}
\end{equation}
\begin{equation}
\ell_{\mathrm{trans}}
=
\operatorname{Huber}
(
s^*\hat{\mathbf{t}}_{ij}
-
\mathbf{t}^{*}_{ij}
).
\label{eq:translation_loss}
\end{equation}
Relative-pose supervision removes global-frame ambiguity and promotes
within-chunk camera consistency.

The resulting depth, confidence, and camera predictions serve as local priors
for TTA. Depth is back-projected from the predicted cameras without a point-map
head, while confidence guides correspondence filtering and weighting; TTA then
corrects residual scale, pose, and coordinate inconsistencies.
\begin{table}[!t]
\centering
\begingroup
\tablefontsize
\setlength{\tabcolsep}{2.4pt}
\renewcommand{\arraystretch}{1.05}
\begin{tabular*}{\columnwidth}{@{\extracolsep{\fill}}lcccc@{}}
\toprule
\multirow{2}{*}{Method}
& \multicolumn{2}{c}{Depth}
& \multicolumn{2}{c}{Camera} \\
\cmidrule(lr){2-3}
\cmidrule(lr){4-5}
& AbsRel$\downarrow$
& SqRel$\downarrow$
& AUC$_5\uparrow$
& AUC$_{30}\uparrow$ \\
\midrule

DA3
& \underline{0.086}
& \underline{0.088}
& \textbf{53.2}
& \underline{77.6} \\

$\pi^3$
& \textbf{0.082}
& 0.229
& 43.3
& 74.9 \\

VGGT
& 0.125
& 0.602
& 43.2
& 68.7 \\

Scal3R
& 0.147
& 0.170
& 34.9
& 67.0 \\

LingBot-Map
& 0.114
& 0.177
& 29.0
& 64.3 \\

GeoWeaver-GPM
& \textbf{0.082}
& \textbf{0.069}
& \underline{49.9}
& \textbf{77.8} \\

\bottomrule
\end{tabular*}
\endgroup

\caption{
GPM evaluation across benchmarks. Camera AUC is reported at
$5^\circ$ and $30^\circ$. Best and second-best results are shown in
bold and underlined, respectively.
}
\label{tab:frontend}
\end{table}

\begin{table*}[!t]
\centering
\begingroup
\widetablefontsize
\setlength{\tabcolsep}{1.8pt}
\renewcommand{\arraystretch}{\widetablerowstretch}
\begin{tabular*}{\textwidth}{@{\extracolsep{\fill}}ll*{4}{cc}@{}}
\toprule
\multirow{2}{*}{Type}
& \multirow{2}{*}{Method}
& \multicolumn{2}{c}{T\&T}
& \multicolumn{2}{c}{Mip-NeRF 360}
& \multicolumn{2}{c}{VKITTI 2}
& \multicolumn{2}{c}{Oxford Spires} \\
\cmidrule(lr){3-4}
\cmidrule(lr){5-6}
\cmidrule(lr){7-8}
\cmidrule(lr){9-10}
&
& AUC$\uparrow$ & ATE$\downarrow$
& AUC$\uparrow$ & ATE$\downarrow$
& AUC$\uparrow$ & ATE$\downarrow$
& RRE$\downarrow$ & ATE$\downarrow$ \\
\midrule

\catopt
& MBA~\tabcite{mba}
& 67.7 & \underline{0.053}
& 43.5 & 0.470
& 30.9 & 46.836
& 3.290 & 29.323 \\

\midrule
\catff
& DA3~\tabcite{da3}
& 63.9 & 0.358
& \textbf{66.4} & \underline{0.037}
& 16.9 & 38.867
& 9.961 & 16.640 \\

\midrule
\catchunk
& VGGT-Long~\tabcite{vggtlong}
& 52.3 & 11.472
& 55.4 & 0.211
& 66.1 & 2.631
& 18.983 & 12.380 \\

\catchunk
& Scal3R~\tabcite{scal3r}
& 58.1 & 0.338
& 37.8 & 0.194
& \underline{67.4} & \textbf{0.608}
& 6.840 & \underline{5.565} \\

\catchunk
& LoGeR~\tabcite{loger}
& 22.8 & 0.757
& 21.1 & 0.455
& 0.0 & 36.183
& 8.100 & \textbf{3.985} \\

\midrule
\catstream
& LingBot-Map~\tabcite{lingbotmap}
& 37.7 & 0.361
& 12.2 & 0.339
& 0.0 & 4.166
& \underline{1.335} & 6.070 \\

\midrule
\cathybrid
& AMB3R~\tabcite{amb3r}
& \textbf{76.9} & 0.261
& 65.0 & 0.186
& \textbf{71.96} & 1.931
& 8.655 & 7.687 \\

\cathybrid
& GeoWeaver
& \underline{72.9} & \textbf{0.023}
& \textbf{66.4} & \textbf{0.036}
& 64.6 & \underline{0.627}
& \textbf{0.575} & \underline{4.372} \\

\bottomrule
\end{tabular*}
\endgroup
\caption{
Average trajectory results across four long-sequence benchmarks. T\&T,
Mip-NeRF~360, VKITTI~2, and Oxford Spires contain 6, 9, 5, and 4 evaluated
scenes or sequences, respectively, spanning object-, room-, road-, and
urban-scale reconstruction. AUC denotes AUC@3$^\circ$ and is reported in
percentage; RRE is in degrees and ATE is in metres. Best and second-best
results are shown in bold and underlined, respectively. Type abbreviations are Opt.
(optimization-based), FF (feed-forward), Chunk-FF (chunk-wise feed-forward),
Stream (streaming), and Hybrid.
}
\label{tab:main_results}
\end{table*}

\begin{table*}[!t]
\centering
\begingroup
\densetablefontsize
\setlength{\tabcolsep}{0.7pt}
\renewcommand{\arraystretch}{\widetablerowstretch}
\begin{tabular*}{\textwidth}{@{\extracolsep{\fill}}ll*{7}{cc}@{}}
\toprule
Type & Method
& \multicolumn{2}{c}{Barn}
& \multicolumn{2}{c}{Caterpillar}
& \multicolumn{2}{c}{Courthouse}
& \multicolumn{2}{c}{Ignatius}
& \multicolumn{2}{c}{Meetingroom}
& \multicolumn{2}{c}{Truck}
& \multicolumn{2}{c}{Average} \\
\cmidrule(lr){3-4} \cmidrule(lr){5-6} \cmidrule(lr){7-8}
\cmidrule(lr){9-10} \cmidrule(lr){11-12} \cmidrule(lr){13-14}
\cmidrule(lr){15-16}
& & AUC$\uparrow$ & ATE$\downarrow$
  & AUC$\uparrow$ & ATE$\downarrow$
  & AUC$\uparrow$ & ATE$\downarrow$
  & AUC$\uparrow$ & ATE$\downarrow$
  & AUC$\uparrow$ & ATE$\downarrow$
  & AUC$\uparrow$ & ATE$\downarrow$
  & AUC$\uparrow$ & ATE$\downarrow$ \\
\midrule
\catopt & MBA~\tabcite{mba}
& 39.3 & 0.144 & 56.5 & 0.026 & 58.9 & 0.069 & 78.6 & 0.038 & 85.4 & 0.025 & 87.7 & 0.017 & 67.7 & 0.053 \\
\midrule
\catff & DA3~\tabcite{da3}
& \ranktwo{69.5} & \rankone{0.039} & \rankthree{63.7} & \rankone{0.023} & 32.4 & 2.000 & 67.4 & \rankthree{0.047} & \ranktwo{76.9} & \ranktwo{0.021} & 73.3 & \ranktwo{0.021} & \rankthree{63.9} & 0.358 \\
\midrule
\catchunk & VGGT-Long~\tabcite{vggtlong}
& \rankthree{65.5} & 0.116 & 46.1 & 0.269 & \ranktwo{51.5} & 2.601 & \rankthree{76.7} & 0.263 & 62.1 & 0.129 & \rankthree{77.5} & 0.067 & 52.3 & 11.472 \\
\catchunk & Scal3R~\tabcite{scal3r}
& 59.9 & 0.053 & 44.6 & 0.158 & \rankthree{43.8} & \rankthree{1.656} & 59.9 & 0.076 & \rankthree{71.3} & \rankthree{0.039} & 69.4 & \rankthree{0.046} & 58.1 & \rankthree{0.338} \\
\catchunk & LoGeR~\tabcite{loger}
& 17.8 & 0.814 & 33.7 & 0.229 & 9.9 & 2.669 & 17.7 & 0.374 & 19.2 & 0.320 & 37.4 & 0.133 & 22.8 & 0.757 \\
\midrule
\catstream & LingBot-Map~\tabcite{lingbotmap}
& 38.1 & 0.063 & 36.9 & 0.126 & 13.9 & 1.695 & 44.7 & 0.107 & 51.9 & 0.075 & 40.6 & 0.100 & 37.7 & 0.361 \\
\midrule
\cathybrid & AMB3R~\tabcite{amb3r}
& \rankone{88.3} & \ranktwo{0.046} & \rankone{87.4} & \rankthree{0.046} & 31.5 & \ranktwo{1.329} & \rankone{94.3} & \ranktwo{0.022} & 65.0 & 0.111 & \rankone{94.9} & \rankone{0.016} & \rankone{76.9} & \ranktwo{0.261} \\
\cathybrid & \ourscell{GeoWeaver}
& \oursval{52.8} & \oursrankthree{0.047} & \oursranktwo{72.9} & \oursranktwo{0.029} & \oursrankone{78.4} & \oursrankone{0.011} & \oursranktwo{77.0} & \oursrankone{0.016} & \oursrankone{78.1} & \oursrankone{0.019} & \oursranktwo{78.2} & \oursranktwo{0.021} & \oursranktwo{72.9} & \oursrankone{0.023} \\
\bottomrule
\end{tabular*}
\endgroup
\caption{Per-scene results on Tanks and Temples. AUC@3$^\circ$ is reported in
percentage and ATE in metres. Best and second-best results are shown in bold
and underlined, respectively.}
\label{tab:tnt}
\end{table*}

\begin{table*}[!t]
\centering
\begingroup
\widetablefontsize
\setlength{\tabcolsep}{1.3pt}
\renewcommand{\arraystretch}{\widetablerowstretch}
\begin{tabular*}{\textwidth}{@{\extracolsep{\fill}}ll*{5}{cc}@{}}
\multicolumn{12}{c}{} \\
\toprule
Type & Method
& \multicolumn{2}{c}{Bicycle}
& \multicolumn{2}{c}{Bonsai}
& \multicolumn{2}{c}{Counter}
& \multicolumn{2}{c}{Flowers}
& \multicolumn{2}{c}{Garden} \\
\cmidrule(lr){3-4} \cmidrule(lr){5-6} \cmidrule(lr){7-8}
\cmidrule(lr){9-10} \cmidrule(lr){11-12}
& & AUC$\uparrow$ & ATE$\downarrow$
  & AUC$\uparrow$ & ATE$\downarrow$
  & AUC$\uparrow$ & ATE$\downarrow$
  & AUC$\uparrow$ & ATE$\downarrow$
  & AUC$\uparrow$ & ATE$\downarrow$ \\
\midrule
\catopt & MBA~\tabcite{mba}
& 79.1 & 0.031 & 78.2 & 0.020 & 84.4 & 0.016 & 8.7 & 0.019 & 5.6 & 3.430 \\
\midrule
\catff & DA3~\tabcite{da3}
& \ranktwo{63.1} & \ranktwo{0.054} & \rankone{76.4} & \rankone{0.027}
& \ranktwo{77.1} & \rankone{0.016} & \ranktwo{72.9} & \ranktwo{0.028}
& 64.4 & \rankthree{0.039} \\
\midrule
\catchunk & VGGT-Long~\tabcite{vggtlong}
& \rankthree{62.9} & \rankthree{0.117} & \rankthree{33.1} & 0.249
& 62.8 & 0.086 & \rankthree{41.9} & 0.525 & \rankthree{78.2} & 0.062 \\
\catchunk & Scal3R~\tabcite{scal3r}
& 42.9 & 0.120 & 27.5 & 0.170 & \rankthree{63.7} & \rankthree{0.033}
& 5.1 & 0.570 & 65.2 & 0.059 \\
\catchunk & LoGeR~\tabcite{loger}
& 34.7 & 0.143 & 9.4 & 0.217 & 25.7 & 0.105 & 6.7 & 2.051 & 28.7 & 0.285 \\
\midrule
\catstream & LingBot-Map~\tabcite{lingbotmap}
& 4.5 & 0.319 & 6.4 & 0.256 & 26.3 & 0.119 & 0.1 & 1.069 & 17.0 & 0.145 \\
\midrule
\cathybrid & AMB3R~\tabcite{amb3r}
& \rankone{85.0} & 0.128 & 28.7 & \rankthree{0.100} & 56.5 & 0.056
& \rankone{85.0} & \rankthree{0.035} & \rankone{89.4} & \ranktwo{0.038} \\
\cathybrid & \ourscell{GeoWeaver}
& \oursval{59.7} & \oursrankone{0.041} & \oursranktwo{48.2} & \oursranktwo{0.067}
& \oursrankone{77.2} & \oursranktwo{0.028} & \oursranktwo{72.9} & \oursrankone{0.024}
& \oursranktwo{85.3} & \oursrankone{0.009} \\
\bottomrule
\end{tabular*}

\vspace{3pt}

\begin{tabular*}{\textwidth}{@{\extracolsep{\fill}}ll*{5}{cc}@{}}
\multicolumn{12}{c}{} \\
\toprule
Type & Method
& \multicolumn{2}{c}{Kitchen}
& \multicolumn{2}{c}{Room}
& \multicolumn{2}{c}{Stump}
& \multicolumn{2}{c}{Treehill}
& \multicolumn{2}{c}{Average} \\
\cmidrule(lr){3-4} \cmidrule(lr){5-6} \cmidrule(lr){7-8}
\cmidrule(lr){9-10} \cmidrule(lr){11-12}
& & AUC$\uparrow$ & ATE$\downarrow$
  & AUC$\uparrow$ & ATE$\downarrow$
  & AUC$\uparrow$ & ATE$\downarrow$
  & AUC$\uparrow$ & ATE$\downarrow$
  & AUC$\uparrow$ & ATE$\downarrow$ \\
\midrule
\catopt & MBA~\tabcite{mba}
& 18.4 & 0.343 & 36.4 & 0.084 & 45.5 & 0.101 & 35.2 & 0.190 & 43.5 & 0.470 \\
\midrule
\catff & DA3~\tabcite{da3}
& \ranktwo{77.1} & \rankone{0.015} & \rankone{55.7} & \rankone{0.029}
& \ranktwo{53.9} & \ranktwo{0.080} & 56.9 & \rankthree{0.044}
& \rankone{66.4} & \ranktwo{0.037} \\
\midrule
\catchunk & VGGT-Long~\tabcite{vggtlong}
& \rankthree{65.3} & 0.081 & \rankthree{48.5} & 0.112 & 36.0 & 0.550
& \rankthree{69.5} & 0.119 & \rankthree{55.4} & 0.211 \\
\catchunk & Scal3R~\tabcite{scal3r}
& 52.9 & \rankthree{0.058} & 40.2 & \rankthree{0.093} & 8.2 & \rankthree{0.507}
& 34.8 & 0.136 & 37.8 & 0.194 \\
\catchunk & LoGeR~\tabcite{loger}
& 31.1 & 0.276 & 17.1 & 0.250 & 13.5 & 0.587 & 24.1 & 0.181 & 21.1 & 0.455 \\
\midrule
\catstream & LingBot-Map~\tabcite{lingbotmap}
& 17.8 & 0.155 & 30.6 & 0.114 & 0.2 & 0.566 & 6.9 & 0.308 & 12.2 & 0.339 \\
\midrule
\cathybrid & AMB3R~\tabcite{amb3r}
& \rankone{77.9} & 0.064 & 33.5 & 0.095 & \rankthree{44.5} & 1.119
& \rankone{84.9} & \ranktwo{0.040} & \rankthree{65.0} & \rankthree{0.186} \\
\cathybrid & \ourscell{GeoWeaver}
& \oursval{57.9} & \oursranktwo{0.052} & \oursranktwo{54.6} & \oursranktwo{0.049}
& \oursrankone{69.9} & \oursrankone{0.029} & \oursranktwo{69.7} & \oursrankone{0.027}
& \oursrankone{66.4} & \oursrankone{0.036} \\
\bottomrule
\end{tabular*}
\endgroup
\caption{Per-scene results on Mip-NeRF 360, split into two panels for
readability. AUC@3$^\circ$ is reported in percentage and ATE in metres. Best
and second-best results are shown in bold and underlined, respectively.}
\label{tab:mip360}
\end{table*}

\begin{table*}[!t]
\centering
\begingroup
\densetablefontsize
\setlength{\tabcolsep}{0.9pt}
\renewcommand{\arraystretch}{\widetablerowstretch}
\begin{tabular*}{\textwidth}{@{\extracolsep{\fill}}ll*{6}{cc}@{}}
\toprule
Type & Method
& \multicolumn{2}{c}{Scene01}
& \multicolumn{2}{c}{Scene02}
& \multicolumn{2}{c}{Scene06}
& \multicolumn{2}{c}{Scene18}
& \multicolumn{2}{c}{Scene20}
& \multicolumn{2}{c}{Average} \\
\cmidrule(lr){3-4} \cmidrule(lr){5-6} \cmidrule(lr){7-8}
\cmidrule(lr){9-10} \cmidrule(lr){11-12} \cmidrule(lr){13-14}
& & AUC$\uparrow$ & ATE$\downarrow$
  & AUC$\uparrow$ & ATE$\downarrow$
  & AUC$\uparrow$ & ATE$\downarrow$
  & AUC$\uparrow$ & ATE$\downarrow$
  & AUC$\uparrow$ & ATE$\downarrow$
  & AUC$\uparrow$ & ATE$\downarrow$ \\
\midrule
\catopt & MBA~\tabcite{mba}
& 36.5 & 39.340 & 23.4 & 1.240 & 36.3 & 0.260 & 52.1 & 18.300 & 6.4 & 175.040 & 30.9 & 46.836 \\
\midrule
\catff & DA3~\tabcite{da3}
& 3.9 & 22.726 & 45.3 & \rankthree{0.151} & 26.1 & 0.075 & 4.6 & 12.061 & 4.4 & 159.321 & 16.9 & 38.867 \\
\midrule
\catchunk & VGGT-Long~\tabcite{vggtlong}
& 56.3 & \rankthree{0.762} & 72.0 & 0.723 & \rankone{58.0} & 0.365 & \ranktwo{85.2} & 1.651 & \ranktwo{59.3} & 9.655 & \rankthree{66.1} & 2.631 \\
\catchunk & Scal3R~\tabcite{scal3r}
& \rankone{81.3} & \rankone{0.376} & \rankthree{74.5} & \rankone{0.038} & \rankthree{42.4} & \ranktwo{0.031} & \rankthree{83.3} & \rankone{0.165} & 55.6 & \ranktwo{2.430} & \ranktwo{67.4} & \rankone{0.608} \\
\catchunk & LoGeR~\tabcite{loger}
& 0.0 & 73.801 & 0.0 & 0.661 & 0.1 & 2.068 & 0.0 & 1.898 & 0.0 & 102.489 & 0.0 & 36.183 \\
\midrule
\catstream & LingBot-Map~\tabcite{lingbotmap}
& 0.0 & 3.170 & 0.0 & 0.560 & 0.0 & 1.010 & 0.0 & 1.170 & 0.0 & 14.920 & 0.0 & 4.166 \\
\midrule
\cathybrid & AMB3R~\tabcite{amb3r}
& \ranktwo{68.4} & 3.037 & \rankone{83.9} & 0.165 & \ranktwo{49.0} & \rankthree{0.040} & \rankone{99.3} & \rankthree{0.534} & \rankthree{59.2} & \rankthree{5.880} & \rankone{71.96} & \rankthree{1.931} \\
\cathybrid & \ourscell{GeoWeaver}
& \oursrankthree{59.8} & \oursranktwo{0.705} & \oursranktwo{76.2} & \oursranktwo{0.087} & \oursval{37.5} & \oursrankone{0.020} & \oursrankthree{83.3} & \oursranktwo{0.473} & \oursrankone{66.3} & \oursrankone{1.851} & \oursval{64.6} & \oursranktwo{0.627} \\
\bottomrule
\end{tabular*}
\endgroup
\caption{Per-scene results on Virtual KITTI 2. AUC@3$^\circ$ is reported in
percentage and ATE in metres. Best and second-best results are shown in bold
and underlined, respectively.}
\label{tab:vkitti2}
\end{table*}

\begin{table*}[!t]
\centering
\begingroup
\widetablefontsize
\setlength{\tabcolsep}{1.2pt}
\renewcommand{\arraystretch}{\widetablerowstretch}
\begin{tabular*}{\textwidth}{@{\extracolsep{\fill}}ll*{5}{cc}@{}}
\toprule
Type & Method
& \multicolumn{2}{c}{keble-04}
& \multicolumn{2}{c}{observatory-01}
& \multicolumn{2}{c}{blenheim-05}
& \multicolumn{2}{c}{christ-church-02}
& \multicolumn{2}{c}{Average} \\
\cmidrule(lr){3-4} \cmidrule(lr){5-6} \cmidrule(lr){7-8}
\cmidrule(lr){9-10} \cmidrule(lr){11-12}
& & RRE$\downarrow$ & ATE$\downarrow$
  & RRE$\downarrow$ & ATE$\downarrow$
  & RRE$\downarrow$ & ATE$\downarrow$
  & RRE$\downarrow$ & ATE$\downarrow$
  & RRE$\downarrow$ & ATE$\downarrow$ \\
\midrule
\catopt & MBA~\tabcite{mba}
& 4.250 & 35.240 & 1.270 & 23.710 & 1.810 & 38.260 & 5.830 & 20.080 & 3.290 & 29.323 \\
\midrule
\catff & DA3~\tabcite{da3}
& 14.442 & 22.664 & 7.087 & 8.365 & \rankthree{2.123} & \ranktwo{1.538} & 16.193 & 33.992 & 9.961 & 16.640 \\
\midrule
\catchunk & VGGT-Long~\tabcite{vggtlong}
& 16.850 & 13.350 & 14.990 & 4.250 & 18.480 & 11.240 & 25.610 & 20.680 & 18.983 & 12.380 \\
\catchunk & Scal3R~\tabcite{scal3r}
& \rankthree{7.830} & \ranktwo{2.130} & \rankthree{5.270} & \ranktwo{1.570} & 7.430 & \rankthree{2.560} & 6.830 & 16.000 & \rankthree{6.840} & \rankthree{5.565} \\
\catchunk & LoGeR~\tabcite{loger}
& 11.000 & \rankthree{4.134} & 8.000 & 5.632 & 9.800 & 2.621 & \rankthree{3.700} & \rankone{3.553} & 8.100 & \rankone{3.985} \\
\midrule
\catstream & LingBot-Map~\tabcite{lingbotmap}
& \ranktwo{2.090} & 8.870 & \ranktwo{0.970} & 6.960 & \ranktwo{1.040} & 4.570 & \ranktwo{1.240} & \ranktwo{3.880} & \ranktwo{1.335} & 6.070 \\
\midrule
\cathybrid & AMB3R~\tabcite{amb3r}
& 8.375 & 4.940 & 5.368 & \rankone{1.220} & 10.740 & 11.830 & 10.140 & \rankthree{12.760} & 8.655 & 7.687 \\
\cathybrid & \ourscell{GeoWeaver}
& \oursrankone{0.432} & \oursrankone{1.430} & \oursrankone{0.360} & \oursrankthree{1.820} & \oursrankone{0.330} & \oursrankone{1.430} & \oursrankone{1.180} & \oursval{12.810} & \oursrankone{0.575} & \oursranktwo{4.372} \\
\bottomrule
\end{tabular*}
\endgroup
\caption{Per-scene results on Oxford Spires. RRE is reported in degrees and
ATE in metres. Best and second-best results are shown in bold and underlined,
respectively.}
\label{tab:oxford}
\end{table*}
\subsection{Test-Time Adaptation}
\label{sec:hierarchical_assembly}

Test-Time Adaptation assembles independently predicted chunks into a globally
consistent reconstruction through three stages with increasing degrees of
freedom. Stages~1 and~2 optimize only chunk-level $\mathrm{Sim}(3)$
transformations while keeping within-chunk poses, depths, and intrinsics fixed.
Stage~3 further refines frame poses, affine depth, and camera-group focal
corrections.

Dense correspondences connect GPM priors to the sequence-specific TTA
objective. We construct a view graph
$\mathcal{H}=(\mathcal{V},\mathcal{E})$ with local temporal and long-range
co-visible edges. Temporal edges connect neighboring frames and adjacent
chunks, SALAD~\cite{salad} retrieves non-local candidates, and
GEP~\cite{gep} selects a compact connected subset with distributed
long-range constraints.

This provides broad scene coverage without matching all
$\mathcal{O}(|\mathcal{V}|^2)$ frame pairs. For each selected edge
$(i,j)\in\mathcal{E}$, RoMa~\cite{roma} extracts dense correspondences
$\mathcal{M}_{ij}$ and matching confidences.

Correspondences are progressively routed through TTA: adjacent cross-chunk
matches initialize neighboring chunk coordinates, cross-chunk and long-range
matches support global chunk alignment, and the complete graph supports
frame-level refinement. Thus, global constraints are introduced only after a
stable initialization.

For a correspondence
$m=(\mathbf{u}_i,\mathbf{u}_j)$, we combine matcher confidence $s_m$ with GPM
confidence
$\gamma_i=\hat{C}_i(\mathbf{u}_i)$ and
$\gamma_j=\hat{C}_j(\mathbf{u}_j)$:
\begin{equation}
w_m
=
s_m\sqrt{\gamma_i\gamma_j},
\qquad
p(m\mid i,j)
=
\frac{w_m}
{\sum_{m'\in\mathcal{M}_{ij}}w_{m'}} .
\label{eq:match_weight}
\end{equation}
These weights guide confidence-aware sampling and residual aggregation. The 2D
term requires a valid correspondence, while the 3D term additionally requires
valid positive depths in both views.

\subsubsection{Stage 1: Sequential Initialization}
\label{sec:stage1}

The first stage constructs an initial global layout from adjacent chunks. For
each neighboring chunk pair, matched pixels are back-projected using their
predicted depths to form corresponding 3D point sets. After confidence and
geometric-consistency filtering, a relative $\mathrm{Sim}(3)$ transformation
is estimated between the two chunks. The adjacent transformations are then
composed sequentially to place all chunks in a common coordinate system.

Only chunk-level transformations are estimated at this stage; the internal
camera poses, depth predictions, and intrinsics of each chunk remain fixed.
Because the initialization relies only on adjacent connections, it may still
accumulate scale and pose drift over long sequences and therefore serves as
the starting point for global chunk alignment.

\subsubsection{Stage 2: Global Chunk Alignment}
\label{sec:stage2}

The second stage performs the main global correction at the chunk level. Each
chunk $\mathcal{C}_k$ is assigned a learnable $\mathrm{Sim}(3)$
transformation:
\begin{equation}
\mathbf{S}_k
=
\begin{bmatrix}
s_k\mathbf{R}_k & \mathbf{t}_k \\
\mathbf{0}^{\top} & 1
\end{bmatrix},
\qquad
s_k=\exp(\alpha_k).
\label{eq:sim3}
\end{equation}
Let $\mathbf{T}_i^{\mathrm{loc}}=[\mathbf{R}_i^{\mathrm{loc}}\mid
\mathbf{t}_i^{\mathrm{loc}}]$ be the GPM
world-to-camera transformation in the local coordinates of observation $i$.
For an observation $i$ belonging to chunk $c(i)$, the corresponding
similarity-valued world-to-camera mapping is
\begin{equation}
\bar{\mathbf{T}}_i
=
\left(
\mathbf{S}_{c(i)}
\left(\mathbf{T}_i^{\mathrm{loc}}\right)^{-1}
\right)^{-1},
\label{eq:aligned_pose}
\end{equation}
which is used to map the fixed local geometry into the global chunk-aligned
coordinate system. It is not itself an $\mathrm{SE}(3)$ pose because it
contains the chunk scale.

The chunk transformations are initialized from Stage~1. Stage~2 jointly
optimizes all non-anchor chunk transformations under the dense 2D
reprojection and 3D consistency objectives, while fixing the first chunk to
remove gauge freedom. The internal frame poses, depth predictions, and
intrinsics remain fixed.

Adjacent cross-chunk correspondences enforce local continuity, whereas
long-range correspondences provide loop-like constraints for correcting
accumulated scale and pose drift. The resulting globally aligned poses and
geometry initialize the frame-level refinement in Stage~3.

\subsubsection{Stage 3: Coarse-to-Fine Global Refinement}
\label{sec:stage3}

Stage~3 refines the globally aligned reconstruction by introducing frame-level
degrees of freedom. For each canonical physical frame $i$, let
$\mathbf{S}_{c(i)}=[s_{c(i)}\mathbf{R}_{c(i)}\mid
\mathbf{t}_{c(i)}]$. We convert the Stage~2 similarity mapping into an
$\mathrm{SE}(3)$ initialization and place depth in the same global scale:
\begin{equation}
\begin{aligned}
\mathbf{T}_i^0
&=
\Bigl[
\mathbf{R}_i^{\mathrm{loc}}\mathbf{R}_{c(i)}^{\top}
\ \Big|\ 
s_{c(i)}\mathbf{t}_i^{\mathrm{loc}}
\\
&\qquad\quad
-\mathbf{R}_i^{\mathrm{loc}}\mathbf{R}_{c(i)}^{\top}
\mathbf{t}_{c(i)}
\Bigr],
\qquad
d_i^0(\mathbf{u})
=
s_{c(i)}\hat{D}_i(\mathbf{u}).
\end{aligned}
\label{eq:stage3_initialization}
\end{equation}

We optimize the initialized world-to-camera pose
$\mathbf{T}_i\in\mathrm{SE}(3)$ using a 6D rotation representation
$\mathbf{r}_i$ and translation $\mathbf{t}_i$. The globally scaled depth is
then corrected using a frame-wise affine model:
\begin{equation}
\tilde{d}_i(\mathbf{u})
=
a_i d_i^0(\mathbf{u})+b_i .
\label{eq:affine_depth}
\end{equation}

To avoid independently fitting focal corrections for every frame, focal
corrections are shared by frames captured with the same physical camera. Let
$g(i)$ denote the camera group of frame $i$. We define
\begin{equation}
\begin{aligned}
f_{x,i}
&=
\hat{f}_{x,i}+\Delta f_{x,g(i)},\\
f_{y,i}
&=
\hat{f}_{y,i}+\Delta f_{y,g(i)}.
\end{aligned}
\label{eq:shared_intrinsics}
\end{equation}
where $\hat{f}_{x,i}$ and $\hat{f}_{y,i}$ are the initial focal estimates.
The active variables in Stage~3 are
\begin{equation}
\Theta_i
=
\left\{
\mathbf{r}_i,
\mathbf{t}_i,
a_i,
b_i
\right\},
\qquad
\Phi_g
=
\left\{
\Delta f_{x,g},
\Delta f_{y,g}
\right\}.
\label{eq:variables}
\end{equation}

The depth and focal corrections are regularized toward their initialization:
\begin{equation}
\begin{aligned}
\mathcal{L}_{\mathrm{reg}}
={}&
\lambda_a\sum_i(a_i-1)^2
+
\lambda_b\sum_i b_i^2\\
&+
\lambda_f\sum_g
\left(
\Delta f_{x,g}^2+
\Delta f_{y,g}^2
\right).
\end{aligned}
\label{eq:parameter_regularization}
\end{equation}

Following Marginalized Bundle Adjustment~\cite{mba}, Stage~3 first optimizes
view-centered subgraphs and then refines the complete selected view graph.
This stage jointly updates camera poses, affine-depth parameters, and optional
shared focal corrections to improve local reprojection accuracy and 3D
consistency. Adaptation terminates once the median reprojection error
stabilizes.

\subsection{Unified Correspondence-Based Objective}
\label{sec:objective}

Stages~2 and~3 are driven by a unified correspondence-based objective; Stage~1
uses robust pairwise $\mathrm{Sim}(3)$ estimation only for initialization.
Given a dense correspondence
$m=(\mathbf{u}_i,\mathbf{u}_j)$ between views $i$ and $j$, the source pixel is
lifted into the source-camera coordinate system using its adjusted depth:
\begin{equation}
\mathbf{X}_i
=
\tilde{d}_i(\mathbf{u}_i)
\mathbf{K}_i^{-1}
\tilde{\mathbf{u}}_i ,
\label{eq:lift}
\end{equation}
where $\tilde{\mathbf{u}}_i$ is the homogeneous pixel coordinate,
$\tilde{d}_i$ is the adjusted depth, and $\mathbf{K}_i$ is the intrinsic
matrix. In Stage~2, the residuals below use
$\bar{\mathbf{T}}_i$ in place of $\mathbf{T}_i$ and the fixed local depth
$\hat{D}_i$ in place of $\tilde{d}_i$; in Stage~3, they use the
$\mathrm{SE}(3)$ poses and affine-corrected depths defined above.
Let $\tilde{\mathbf{X}}_i=(\mathbf{X}_i^{\top},1)^{\top}$ denote the
homogeneous lifting of $\mathbf{X}_i$, and let
$\operatorname{dehom}(\cdot)$ discard the homogeneous coordinate.

Since $\mathbf{T}_i$ denotes a world-to-camera transformation, the 2D
reprojection residual is
\begin{equation}
r_{\mathrm{2D},m}
=
\left\|
\pi\!\left(
\mathbf{K}_j
\begin{bmatrix}\mathbf{I}_3 & \mathbf{0}\end{bmatrix}
\mathbf{T}_j
\mathbf{T}_i^{-1}
\tilde{\mathbf{X}}_i
\right)
-
\mathbf{u}_j
\right\|_2 ,
\label{eq:residual_2d}
\end{equation}
where $\pi(\cdot)$ denotes perspective projection. When valid positive depths
are available in both views, we additionally define the world-space residual
\begin{equation}
r_{\mathrm{3D},m}
=
\left\|
\operatorname{dehom}\!\left(\mathbf{T}_i^{-1}\tilde{\mathbf{X}}_i\right)
-
\operatorname{dehom}\!\left(\mathbf{T}_j^{-1}\tilde{\mathbf{X}}_j\right)
\right\|_2 ,
\label{eq:residual_3d}
\end{equation}
where $\mathbf{X}_j$ is lifted from the matched target pixel. The 2D residual
is defined for every valid correspondence, whereas the 3D residual requires
valid depths in both views.

Dense matching and predicted depth provide large sets of 2D and 3D residuals,
which can be regarded as samples from empirical error distributions rather
than isolated measurements. This motivates a distribution-level objective
that moves a larger fraction of correspondences toward the low-residual
region. Since dense wide-baseline matches also contain structured outliers, we
use a smooth multi-threshold CDF instead of minimizing the mean residual or
using a single hard inlier threshold.

Following the threshold-marginalization motivation of
MBA~\cite{mba}, for optimization stage $s$ and residual type
$q\in\{\mathrm{2D},\mathrm{3D}\}$, we first apply a stage-dependent residual
mapping:
\begin{equation}
e_{q,m}^{(s)}
=
\rho_s\!\left(r_{q,m}\right).
\label{eq:residual_schedule}
\end{equation}
We use the logistic function
\begin{equation}
\sigma(z)
=
\frac{1}{1+\exp(-z)}
\label{eq:logistic_smoothing}
\end{equation}
as a differentiable approximation to the hard threshold indicator. The
thresholds are uniformly distributed as
\begin{equation}
\begin{aligned}
\mathcal{T}_{q}^{(s)}
&=
\left\{
\tau_{q,\ell}^{(s)}
\right\}_{\ell=1}^{L_q},\\
\tau_{q,\ell}^{(s)}
&=
\frac{\ell}{L_q}
\tau_{q,\max}^{(s)},\\
\beta_q^{(s)}
&=
\kappa_q^{(s)}
\frac{\tau_{q,\max}^{(s)}}{L_q},
\end{aligned}
\label{eq:threshold_grid}
\end{equation}
where $L_q$ is the number of thresholds,
$\tau_{q,\max}^{(s)}$ is the maximum threshold, and
$\kappa_q^{(s)}$ controls the smoothing bandwidth.

The confidence-weighted empirical CDF is
\begin{equation}
F_{q}^{(s)}(\tau)
=
\frac{
\sum_m
w_m v_{q,m}
\sigma\!\left(
\frac{\tau-e_{q,m}^{(s)}}{\beta_q^{(s)}}
\right)
}{
\sum_m w_m v_{q,m}+\epsilon
},
\label{eq:weighted_cdf}
\end{equation}
where $w_m$ is the correspondence confidence,
$v_{q,m}$ is the validity indicator for residual type $q$, and $\epsilon$
ensures numerical stability. The corresponding CDF loss is
\begin{equation}
\mathcal{L}_{\mathrm{CDF},q}^{(s)}
=
\frac{1}{L_q}
\sum_{\ell=1}^{L_q}
\left[
1-
F_q^{(s)}
\left(
\tau_{q,\ell}^{(s)}
\right)
\right].
\label{eq:cdf_loss}
\end{equation}
Minimizing this loss increases the weighted proportion of correspondences in
the low-residual region across multiple thresholds, making the objective less
sensitive to extreme outliers and threshold selection.

The complete objective at stage $s$ is
\begin{equation}
\mathcal{L}^{(s)}
=
\mathcal{L}_{\mathrm{CDF},\mathrm{2D}}^{(s)}
+
\lambda_{\mathrm{3D}}
\mathcal{L}_{\mathrm{CDF},\mathrm{3D}}^{(s)}
+
\mathcal{L}_{\mathrm{reg}},
\label{eq:final_objective}
\end{equation}
where $\lambda_{\mathrm{3D}}$ balances the 2D and 3D terms and
$\mathcal{L}_{\mathrm{reg}}$ regularizes the optimized variables. During
Stage~2, $\mathcal{L}_{\mathrm{reg}}=0$ because only chunk similarities are
active; during Stage~3 it is given by Eq.~\ref{eq:parameter_regularization}.
The CDF objective is used throughout the optimization stages of TTA.
Chunk-level adaptation corrects
large-scale similarity inconsistencies, while coarse-to-fine frame-level
adaptation improves local reprojection and 3D consistency. All TTA
settings are shared across datasets and source models, with exact values
reported in the supplementary material below.

\section{Experiments}
\label{sec:experiments}

\subsection{Experimental Setup}

\paragraph{Datasets and metrics.}
We evaluate GeoWeaver on Tanks and Temples~\cite{Knapitsch2017},
Mip-NeRF~360~\cite{barron2022mipnerf360}, Virtual KITTI~2
~\cite{cabon2020virtual}, and Oxford Spires~\cite{oxfordspires}, covering
indoor, outdoor, driving, and large-scale scenes. We report ATE, RRE, and
AUC@3$^\circ$, with ATE computed after $\mathrm{Sim}(3)$ alignment. For the
standalone GPM, we additionally report depth AbsRel and SqRel and camera pose
AUC.

\paragraph{Baselines and protocol.}
We compare with optimization-based, feed-forward, chunk-wise, streaming, and
hybrid methods. Chunk-wise baselines use 60-frame chunks with 30-frame
overlap, whereas GeoWeaver uses only one shared frame, increasing the stride
from 30 to 59 and reducing redundant GPM inference. Despite this minimal
overlap, GeoWeaver maintains strong global consistency and trajectory
accuracy.

\paragraph{Implementation details.}
The GPM builds on Depth Anything~3~\cite{da3} and is trained on mixed indoor,
outdoor, and synthetic data using 2--16-view clips. We use AdamW with separate
encoder and head learning rates for 200K iterations under cosine decay.
Candidate edges include temporal neighbors and SALAD-retrieved non-local
pairs; GEP~\cite{gep} selects a compact connected graph, and
RoMa~\cite{roma} extracts dense correspondences. Retrieval, matching,
weighting, CDF, and TTA settings are shared across datasets and source models.
Exact settings are provided in the supplementary material below.

\subsection{Geometric Prior Model Evaluation}

Table~\ref{tab:frontend} shows that GeoWeaver-GPM is competitive in both depth
and camera estimation, achieving the best SqRel and AUC@30$^\circ$. Its depth,
confidence, and camera predictions therefore provide reliable priors for
long-sequence adaptation.

\subsection{Main Pipeline Evaluation}

Table~\ref{tab:main_results} summarizes results on four long-sequence
benchmarks, with per-scene results provided in the supplement. GeoWeaver
achieves the lowest average ATE on Tanks and Temples and Mip-NeRF~360, the
second-lowest ATE on Virtual KITTI~2 despite one-frame overlap, and the lowest
RRE on Oxford Spires. These results demonstrate improved trajectory accuracy
and long-range rotational consistency.

\begin{figure}[t]
\centering
\includegraphics[
    width=0.95\linewidth,
    keepaspectratio
]{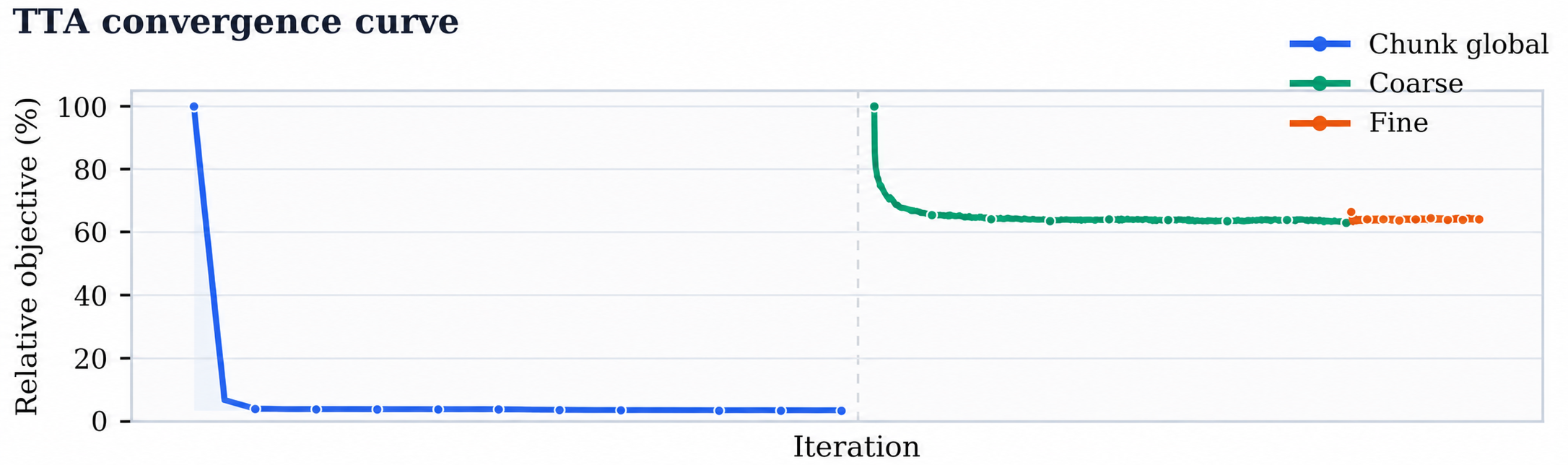}
\caption{
Convergence of GeoWeaver TTA. The objective decreases rapidly during
Stage~2 chunk-level alignment and saturates during Stage~3 refinement.
}
\label{fig:supp_backend_convergence}
\end{figure}

Figure~\ref{fig:qualitative} further shows that competing chunk-wise methods
produce discontinuities or misaligned structures at chunk boundaries, whereas
GeoWeaver preserves local GPM detail within a coherent global reconstruction.
Additional qualitative results, CDF analysis, and convergence curves are
provided in the supplementary material below.

\paragraph{Convergence and efficiency.}
GeoWeaver's TTA reduces the objective by 50\% within the first 100
chunk-alignment iterations and approaches convergence after approximately 440
iterations. Chunk-level alignment removes most pose and scale inconsistencies
before frame-level refinement, while median-error-based stopping avoids
unnecessary updates. GPM inference, edge matching, and residual evaluation are
GPU-parallelizable, supporting efficient long-sequence optimization.
\subsection{Ablation Studies}

We analyze the hierarchical TTA procedure and its compatibility with different
Geometric Prior Models on Tanks and Temples.

\begin{table}[!t]
\centering
\begingroup
\tablefontsize
\setlength{\tabcolsep}{3pt}
\renewcommand{\arraystretch}{1.05}
\begin{tabular*}{\columnwidth}{@{\extracolsep{\fill}}ccccc@{}}
\toprule
Seq. Init.
&
Chunk Align.
&
Frame Refine
&
ATE$\downarrow$
&
AUC$\uparrow$ \\
\midrule

$\checkmark$
& --
& --
& 0.674
& 24.4 \\

$\checkmark$
& $\checkmark$
& --
& 0.075
& 43.8 \\

$\checkmark$
& $\checkmark$
& $\checkmark$
& \textbf{0.023}
& \textbf{72.9} \\

\bottomrule
\end{tabular*}
\endgroup
\caption{
Ablation of the three-stage Test-Time Adaptation procedure on Tanks and
Temples. AUC denotes AUC@3$^\circ$. The full configuration progressively
introduces sequential initialization, chunk-level adaptation, and
frame-level adaptation. Highlighted ablation results are shown in bold.
}
\label{tab:ablation_stages}
\end{table}

\begin{table}[!t]
\centering
\begingroup
\tablefontsize
\setlength{\tabcolsep}{3.6pt}
\renewcommand{\arraystretch}{1.05}
\begin{tabular*}{\columnwidth}{@{\extracolsep{\fill}}lcc@{}}
\toprule
Prior Model
& ATE$\downarrow$
& AUC$\uparrow$ \\
\midrule

Scal3R~\tabcite{scal3r}
& 0.338
& 58.1 \\

Scal3R + TTA
& \textbf{0.033}
& \textbf{63.6} \\

\addlinespace[1pt]

DA3~\tabcite{da3}
& 0.358
& 63.9 \\

DA3 + TTA
& \textbf{0.033}
& \textbf{68.5} \\

\addlinespace[1pt]

GeoWeaver-GPM + TTA
& \textbf{0.025}
& \textbf{71.3} \\

\bottomrule
\end{tabular*}
\endgroup
\caption{
Compatibility of Test-Time Adaptation with different Geometric Prior Models
on Tanks and Temples. AUC denotes AUC@3$^\circ$. The same TTA configuration
improves multiple prior models without source-specific tuning. Highlighted
ablation results are shown in bold.
}
\label{tab:ablation_priors}
\end{table}

\paragraph{Hierarchical adaptation.}
Table~\ref{tab:ablation_stages} separates the role of each TTA stage.
Sequential initialization provides a connected but drift-prone layout.
Chunk-level adaptation then removes most large cross-chunk discrepancies,
yielding the largest ATE reduction. Frame-level adaptation makes a smaller but
decisive correction to camera poses, affine depth, and focal corrections, increasing
AUC@3$^\circ$ from 43.8 to 72.9. These trends support the coarse-to-fine
adaptation order rather than treating all geometric variables as equally
reliable from the start.

\paragraph{Geometric prior model compatibility.}
Applying the same TTA procedure to Scal3R and DA3 substantially reduces ATE and
improves AUC. All source models use the identical retrieval graph, RoMa
matches, adaptive stopping criteria, and TTA hyperparameters. The consistent
gains indicate that TTA exploits geometric information shared across
different Geometric Prior Models rather than relying on source-specific
tuning.

\section{Conclusions and Limitations}
\label{sec:conclusion}

We presented \textbf{GeoWeaver}, a hierarchical framework for long-sequence
3D reconstruction that combines a Geometric Prior Model (GPM) with Test-Time
Adaptation (TTA). The GPM predicts chunk-wise depth, confidence, and cameras;
TTA sequentially initializes the global layout, aligns chunks with
$\mathrm{Sim}(3)$ transformations, and refines frame poses, depth, and focal
calibration. Experiments and ablations validate this coarse-to-fine design
across multiple geometric priors.

GeoWeaver incurs additional matching and adaptation cost over feed-forward
inference, and currently assumes static scenes and fixed per-camera focal
calibration. Extending it to dynamic scenes, rolling shutter, zoom, and
real-time operation remains future work. Post-training the GPM with the TTA
objective may further improve global consistency and reduce test-time
optimization.

\bibliographystyle{plainnat}
\bibliography{references}
\clearpage
\appendix

\twocolumn[
\begin{center}
{\Large\bfseries Supplementary Material}
\end{center}
\vspace{0.8em}
]
\captionsetup[figure]{
  font=small,
  labelfont=bf,
  justification=justified,
  singlelinecheck=false,
  skip=4pt
}

\section{Geometric Prior Model Details}

\subsection{Training Data}

We train the Geometric Prior Model (GPM) on a mixture of real and synthetic multi-view datasets: Hypersim, ScanNet, ScanNet++, MegaSynth, ASE, MVSSynth, Unreal4K, BlendedMVS, DynamicStereo, TartanAir, and HM3D. The mixture is designed to expose the model to complementary scene statistics rather than a single capture domain. It includes real RGB-D scans, high-fidelity synthetic interiors, procedurally generated geometry, wide-baseline multi-view imagery, dynamic content, and challenging camera trajectories. Table~\ref{tab:training_data} summarizes its composition.

For each source, RGB images form the common model input. We convert the available depth, camera intrinsics, camera extrinsics, and reconstructed geometry into the unified local-depth, depth-induced global-geometry, confidence-aware geometry, and relative-pose supervision described in the main paper. Validity masks are inherited from the source annotations so that missing or undefined geometry does not contribute to the loss. Multi-view clips are sampled from views belonging to the same scene or sequence, preserving their calibrated geometric relationships.

The data sources are complementary. Real captures reduce the synthetic-to-real appearance gap, whereas synthetic datasets provide dense and complete supervision that is difficult to obtain from physical sensors. Indoor scans emphasize clutter, occlusion, and room-scale structure; procedural and MVS datasets expand the range of layouts, baselines, and scene scales; and dynamic or navigation-oriented sequences expose the model to nontrivial temporal variation and camera motion. Together, this mixture supports a GPM that must remain stable across indoor, outdoor, synthetic, and long-sequence reconstruction benchmarks.

\subsection{Two-Stage Training Strategy}

We progressively increase input diversity rather than expose the model to all
sequence lengths from the outset. In the first stage, every sample contains
four frames, and we train on eight NVIDIA H20 GPUs for 100K iterations. This
fixed-length stage establishes stable local geometry and camera predictions.

In the second stage, we train for a further 100K iterations with variable-length
clips containing 2--16 frames; each GPU receives at most 16 input frames per
iteration. This stage exposes the model to the view counts and baselines that
arise during chunked inference. The complete procedure therefore uses 200K
iterations on eight H20 GPUs: first to stabilize local predictions, and then
to make those predictions robust to varying chunk configurations.
\subsection{Optimization}

We optimize the model with AdamW and use separate learning rates for the shared encoder and prediction heads. The encoder learning rate is $1\times10^{-6}$, while all task heads use $1\times10^{-5}$. Both stages use a cosine learning-rate schedule (CosineLR). The lower encoder rate preserves the pretrained visual representation, whereas the higher head rate allows the depth, confidence, and camera heads to adapt to the multi-dataset geometric supervision.
\begin{table}[t]
\centering
\caption{Training datasets grouped by data domain.}
\label{tab:training_data}
\suppsingletablestyle
\begin{tabular*}{\columnwidth}{
@{\extracolsep{\fill}}
p{0.30\columnwidth}
p{0.62\columnwidth}
@{}
}
\toprule
\textbf{Data domain} & \textbf{Datasets} \\
\midrule
Real indoor & ScanNet, ScanNet++, HM3D \\
Synthetic indoor & Hypersim, ASE \\
Procedural & MegaSynth \\
Multi-view / stereo & MVSSynth, Unreal4K, BlendedMVS \\
Dynamic / motion & DynamicStereo, TartanAir \\
\bottomrule
\end{tabular*}
\end{table}
\section{Test-Time Adaptation Details}
\label{sec:supp_backend}
This section specifies the fixed TTA settings that assemble GPM priors into a
global reconstruction. TTA applies three stages with increasing degrees of
freedom: sequential initialization, global chunk-level alignment, and
coarse-to-fine frame-level refinement.
\subsection{Optimization Settings}

TTA constructs a candidate graph from temporal neighbors and SALAD-retrieved
non-local pairs. GEP retains a compact connected subset, and RoMa provides the
dense correspondences used in all three stages. Stage~1 performs robust,
confidence-weighted pairwise $\mathrm{Sim}(3)$ estimation on the top $50\%$
most confident correspondences to initialize the chunk layout. In Stage~2, the
first chunk is fixed to remove gauge freedom and the remaining chunk
transformations are optimized with Adam at a learning rate of $10^{-2}$. This
optimization is capped at $5{,}000$ iterations and uses the CDF-based 2D
reprojection objective and a 3D consistency term with weight $2.0$ and maximum
distance $0.8$. The resulting chunk similarities initialize the world-to-camera
poses and globally scaled depths for Stage~3.

Stage~3 frame-level refinement uses Adam with a learning rate of $10^{-4}$.
Coarse refinement operates on view-centered subgraphs using a CDF range of
$15$ pixels, $250$ bins, and a smoothing bandwidth of $2$. Fine refinement
then jointly optimizes the complete selected view graph. Both phases optimize
frame poses and affine depth, with a 3D consistency term of weight $1.0$ and
maximum distance $0.1$. For uncalibrated sequences, focal corrections are
shared by frames from the same physical-camera group; for calibrated sequences,
the initial focal lengths remain fixed.

Table~\ref{tab:supp_backend_hparams} summarizes the default settings shared across all datasets and source priors.
\begin{table}[t]
\centering
\caption{Default hyperparameters of GeoWeaver TTA.}
\label{tab:supp_backend_hparams}
\suppsingletablestyle
\begin{tabular*}{\columnwidth}{
@{\extracolsep{\fill}}
p{0.30\columnwidth}
p{0.43\columnwidth}
c
@{}
}
\toprule
\textbf{Stage} & \textbf{Parameter} & \textbf{Value} \\
\midrule
Stage~1: sequential initialization
& Confidence retention
& $50\%$ \\
\midrule
Stage~2: global chunk alignment
& Learning rate
& $10^{-2}$ \\
& Maximum iterations
& $5{,}000$ \\
& 3D range / weight
& $0.8$ / $2.0$ \\
\midrule
Stage~3: coarse refinement
& CDF range / bins
& $15$ / $250$ \\
& Smoothing bandwidth
& $2$ \\
\midrule
Stage~3: fine refinement
& Learning rate
& $10^{-4}$ \\
& 3D range / weight
& $0.1$ / $1.0$ \\
& Focal correction
& Camera-group shared$^{\dagger}$ \\
\midrule
\shortstack[l]{Confidence- and\\consistency-aware\\sampling (CCAS)}
& Confidence threshold
& $0.5$ \\
& Depth-consistency threshold
& $0.2$ \\
& Maximum samples per pair
& $10{,}000$ \\
\bottomrule
\end{tabular*}
\end{table}

\noindent\footnotesize $^{\dagger}$Enabled only for uncalibrated sequences; calibrated focal lengths remain fixed.\normalsize

\subsection{Confidence- and Consistency-Aware Sampling and Early Stopping}

Confidence- and Consistency-Aware Sampling (CCAS) selects correspondences that
are jointly supported by matcher confidence and predicted geometry. We use a
relative confidence threshold of $0.5$ and a depth-consistency threshold of
$0.2$. When this strict filtering leaves too few constraints to connect the
graph, the depth-consistency threshold is progressively relaxed rather than
discarding the pair. The sampling budget is adjusted to graph size and
available GPU memory, with at most $10{,}000$ correspondences retained per
image pair.

All iterative stages use adaptive stopping based on the median reprojection
error. Terminating once this robust statistic stabilizes avoids spending a
fixed maximum iteration budget on already converged sequences while preserving
additional updates for difficult ones.

\subsection{Residual Distributions}

\begin{figure}[t]
\centering
\includegraphics[
    width=0.95\linewidth,
    keepaspectratio
]{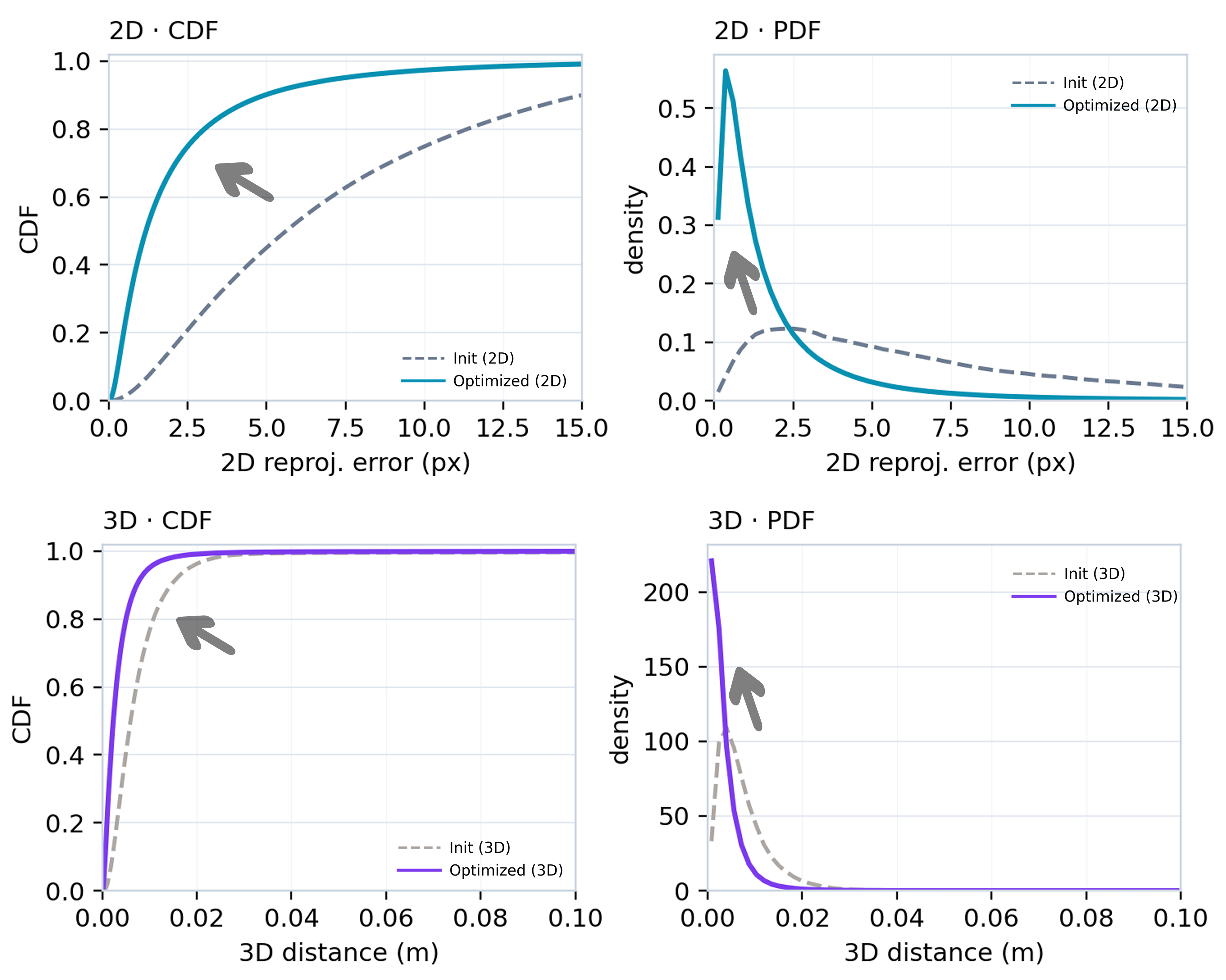}
\caption{Residual distributions before and after optimization. The optimized 2D reprojection and 3D distance residuals concentrate closer to zero, producing steeper CDF curves and sharper PDF peaks.}
\label{fig:supp_objective_cdf}
\end{figure}

Figure~\ref{fig:supp_objective_cdf} makes the effect of the final objective
explicit. For both 2D reprojection errors and 3D metric distances, optimization
moves more high-confidence correspondences into the low-residual regime: the
CDF rises earlier and the PDF concentrates closer to zero. This is the desired
behavior of the CDF-style objective, which rewards improving the inlier
distribution rather than fitting a small set of already easy matches.

\subsection{Convergence and Efficiency}

\begin{figure}[t]
\centering
\includegraphics[
    width=0.95\linewidth,
    keepaspectratio
]{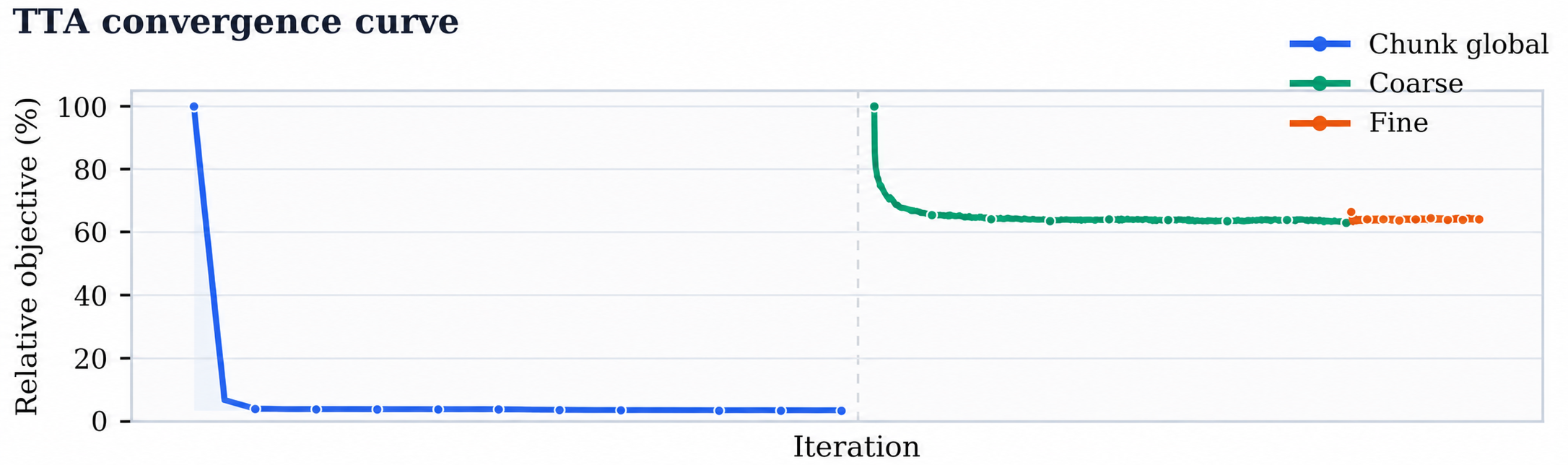}
\caption{Convergence curve of GeoWeaver TTA. The objective drops rapidly during Stage~2 chunk-level alignment and then saturates during Stage~3 refinement.}
\label{fig:supp_backend_convergence_detail}
\end{figure}

Figure~\ref{fig:supp_backend_convergence_detail} demonstrates the rapid convergence
of hierarchical TTA. The objective decreases by 50\% within the first
100 chunk-alignment iterations and approaches convergence after approximately
440 iterations. Since most large-scale scale and pose errors are removed during
this low-dimensional chunk-level optimization, the subsequent coarse and fine
refinement stages require only limited additional updates.

This convergence behavior substantially reduces the effective optimization
budget of TTA. Together with the adaptive stopping criterion, which
terminates each stage once the median reprojection error stabilizes, GeoWeaver
avoids unnecessary iterations on already well-aligned sequences. Moreover,
chunk-wise GPM inference, dense matching over selected graph edges, and
correspondence-residual evaluation can all be executed in parallel on GPUs.
Consequently, TTA achieves efficient global refinement despite
optimizing long sequences and a global view graph.

\section{Dataset and Evaluation Protocol}
\label{sec:supp_evaluation}
This section fixes the alignment and metric conventions used throughout the
paper. In particular, all trajectory metrics use one sequence-level Sim(3)
alignment; no per-frame or per-chunk realignment is applied, so sequence-level drift
remains reflected in the reported metrics.
\subsection{Trajectory Alignment}
For a world-to-camera pose
$\mathbf{T}_i=[\mathbf{R}_i\mid\mathbf{t}_i]$, its camera center is
\begin{equation}
\mathbf{c}_i=-\mathbf{R}_i^{\top}\mathbf{t}_i.
\end{equation}
Because monocular reconstruction is defined up to a global similarity
transformation, we align the predicted camera centers to the reference centers
using Umeyama alignment:
\begin{equation}
(s,\mathbf{R}_a,\mathbf{t}_a)
=
\arg\min_{s,\mathbf{R},\mathbf{t}}
\sum_{i=1}^{N}
\left\|
s\mathbf{R}\hat{\mathbf{c}}_i+\mathbf{t}
-\mathbf{c}_i^{*}
\right\|_2^2 .
\label{eq:supp_sim3_alignment}
\end{equation}
The aligned camera centers and world-to-camera rotations are
\begin{equation}
\bar{\mathbf{c}}_i
=
s\mathbf{R}_a\hat{\mathbf{c}}_i+\mathbf{t}_a,
\qquad
\bar{\mathbf{R}}_i
=
\hat{\mathbf{R}}_i\mathbf{R}_a^{\top}.
\label{eq:supp_aligned_pose}
\end{equation}
A single Sim(3) transformation is estimated for each complete sequence; no
per-frame alignment is performed.
\subsection{Absolute Trajectory Error}
We first compute the translation error of every aligned camera center:
\begin{equation}
e_{t,i}
=
\left\|
\bar{\mathbf{c}}_i-\mathbf{c}_i^{*}
\right\|_2 .
\end{equation}
The ATE reported in the main paper is its root-mean-square value:
\begin{equation}
\mathrm{ATE}
=
\sqrt{
\frac{1}{N}
\sum_{i=1}^{N} e_{t,i}^{2}
}.
\label{eq:supp_ate}
\end{equation}
ATE is measured in metres, and lower is better.
\subsection{Rotation Error}
For each matched frame, we compute the geodesic error between the aligned and
reference rotations:
\begin{equation}
e_{R,i}
=
\frac{180}{\pi}
\arccos\!\left[
\operatorname{clip}\!\left(
\frac{
\operatorname{tr}\!\left(
\bar{\mathbf{R}}_i
(\mathbf{R}_i^{*})^{\top}
\right)-1
}{2},
-1,1
\right)
\right].
\label{eq:supp_rotation_error}
\end{equation}
The RRE values in the main tables correspond to the arithmetic mean
\begin{equation}
\mathrm{RRE}
=
\frac{1}{N}\sum_{i=1}^{N}e_{R,i},
\label{eq:supp_rre}
\end{equation}
reported in degrees. Thus, RRE denotes the mean rotation registration error
after sequence-level Sim(3) alignment rather than consecutive-frame RPE.
\subsection{Pairwise Pose AUC}
Pose AUC follows the IMC2021 evaluation protocol. For every evaluated ordered
pair $(i,j)$, we construct the relative world-to-camera transformations
\begin{equation}
\mathbf{T}_{ij}^{*}
=
\mathbf{T}_{j}^{*}
(\mathbf{T}_{i}^{*})^{-1},
\qquad
\bar{\mathbf{T}}_{ij}
=
\bar{\mathbf{T}}_{j}
\bar{\mathbf{T}}_{i}^{-1}.
\label{eq:supp_relative_pose}
\end{equation}
The pairwise rotation error is
\begin{equation}
e_{R,ij}
=
\frac{180}{\pi}
\arccos\!\left[
\operatorname{clip}\!\left(
\frac{
\operatorname{tr}\!\left(
\mathbf{R}_{ij}^{*}
\bar{\mathbf{R}}_{ij}^{\top}
\right)-1
}{2},
-1,1
\right)
\right].
\end{equation}
Following the IMC protocol, the translation error is invariant to the sign of
the translation direction:
\begin{equation}
e_{t,ij}
=
\frac{180}{\pi}
\arccos\!\left(
\left|
\frac{
(\mathbf{t}_{ij}^{*})^{\top}\bar{\mathbf{t}}_{ij}
}{
\|\mathbf{t}_{ij}^{*}\|_2
\|\bar{\mathbf{t}}_{ij}\|_2
}
\right|
\right).
\label{eq:supp_translation_angle}
\end{equation}
The final pairwise pose error is
\begin{equation}
e_{ij}=\max(e_{R,ij},e_{t,ij}).
\end{equation}
Given the empirical recall curve
\begin{equation}
P(\tau)
=
\frac{1}{|\mathcal{E}_{\mathrm{eval}}|}
\sum_{(i,j)\in\mathcal{E}_{\mathrm{eval}}}
\mathbb{I}[e_{ij}\leq\tau],
\end{equation}
we compute
\begin{equation}
\mathrm{AUC}@\theta
=
\frac{1}{\theta}
\int_{0}^{\theta}P(\tau)\,d\tau.
\label{eq:supp_pose_auc}
\end{equation}
The implementation returns values in $[0,1]$, which are multiplied by $100$ for
presentation in the main tables. We report AUC@3$^\circ$ for the full pipeline.
All ordered pairs with $i\neq j$ are used when their number does not exceed
$20{,}000$; otherwise, $20{,}000$ ordered pairs are sampled without replacement
using a fixed random seed of 42.

\subsection{Depth Metrics}
Standalone GPM depth is evaluated over the valid-pixel set $\Omega$ in
each camera coordinate frame. Given predicted and reference depths
$\hat d(\mathbf{u})$ and $d^{*}(\mathbf{u})$, we report
\begin{align}
\mathrm{AbsRel}
&=
\frac{1}{|\Omega|}
\sum_{\mathbf{u}\in\Omega}
\frac{
|\hat d(\mathbf{u})-d^{*}(\mathbf{u})|
}{
d^{*}(\mathbf{u})
},\\
\mathrm{SqRel}
&=
\frac{1}{|\Omega|}
\sum_{\mathbf{u}\in\Omega}
\frac{
(\hat d(\mathbf{u})-d^{*}(\mathbf{u}))^2
}{
d^{*}(\mathbf{u})
}.
\end{align}
Invalid reference pixels and non-positive predicted depths are excluded.

\onecolumn
\section{Additional Qualitative Results}
\label{sec:supp_qualitative}

We provide additional visualizations across diverse indoor, outdoor, and
large-scale scenes. The following full-width composite figure jointly
visualizes the reconstructed point clouds and estimated camera poses for
multiple sequences. This compact overview facilitates direct inspection of
geometric completeness, trajectory consistency, and long-range drift across
different scene types.

\begin{figure}[!ht]
  \centering
  \includegraphics[
    width=1.08\textwidth,
    height=0.76\textheight,
    keepaspectratio
  ]{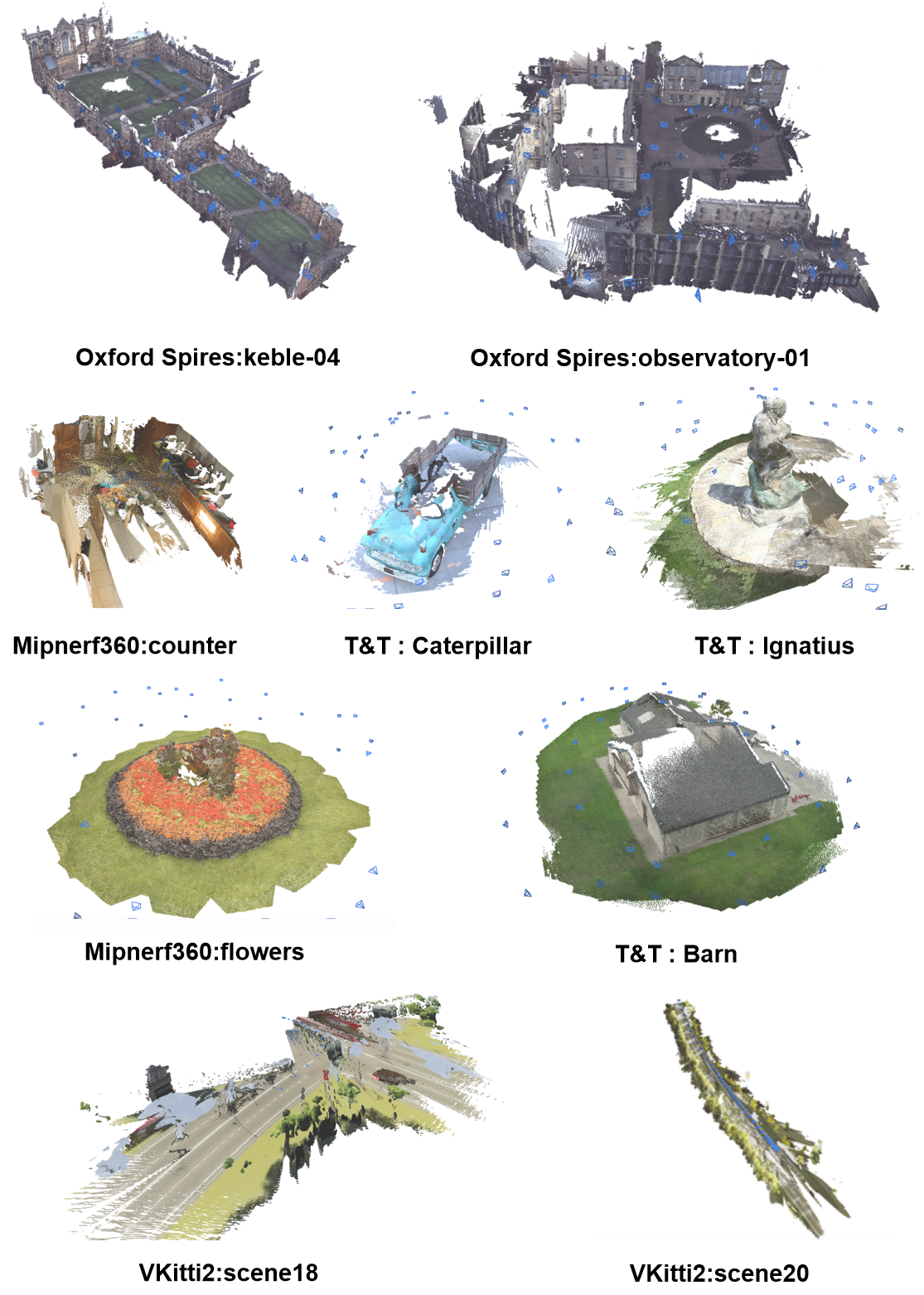}
  \caption{Additional qualitative results across diverse scenes. For each
  scene, we visualize the reconstructed point cloud together with the estimated
  camera poses.}
  \label{fig:supp_qualitative_results}
\end{figure}

\end{document}